\documentclass[11pt]{article}

\usepackage[preprint]{Configs/NAACL/acl}

\usepackage{times}
\usepackage{latexsym}

\usepackage[T1]{fontenc}    

\usepackage[utf8]{inputenc} 

\usepackage{microtype}      

\usepackage{inconsolata}

\usepackage{graphicx}

\usepackage{url}            
\usepackage{booktabs}       
\usepackage{amsfonts}       
\usepackage{nicefrac}       
\usepackage{svg}
\usepackage{amsmath,amssymb}
\usepackage{enumitem}
\usepackage{multirow} 
\usepackage{adjustbox}

\usepackage{xcolor}

\usepackage{Configs/mathcomEd}
\usepackage{Configs/mathstatsEd} 

\usepackage{siunitx}

\usepackage{booktabs,colortbl}

\usepackage{stfloats}        

\definecolor{bestcol}{RGB}{230,245,208} 
\definecolor{secondcol}{RGB}{255,247,188} 
\definecolor{Burgundy}{RGB}{144,0,32}

\newcommand{\eduin}{ \color{ForestGreen}}

\title{
    How Benchmark Graph Construction Shapes Multi-Hop KG Navigation: \\
    A Controlled Audit of PathQuestion and PathQuestion-Large
}

\author{
    Eduin E.~Hernandez$^{1}$\thanks{Corresponding author: \texttt{eduin.ee08@nycu.edu.tw}.}\quad
    Sergio A.~Diaz$^{1}$ \quad
    Luis F.~Garcia$^{2}$ \quad
    Nurassyl Askar$^{2}$ \quad
    Stefano Rini$^{1}$ \\
    $^{1}$ National Yang Ming Chiao Tung University, Hsinchu, Taiwan \\
    $^{2}$ Independent Researcher \\
}

\begin{document}

\maketitle

\begin{abstract}
Multi-hop knowledge-graph question answering (KGQA) benchmarks are often treated as fixed language-and-reasoning tasks, but for navigation-based methods, difficulty also depends on the graph in which those tasks are instantiated.
This makes graph construction a consequential but often entangled part of the navigation task.
To isolate this effect, we audit PathQuestion (PQ) and PathQuestion-Large (PQL) as graph-navigation benchmarks under a controlled intervention: we hold questions, topic entities, released answers, released path annotations, and family-level splits fixed while replacing each released task-filtered graph with the larger source KG distributed with the benchmark.
Source restoration expands the effective navigation search space and graph-consistent answer and path multiplicity while sharply reducing unguided reachability to released answers.
We evaluate whether these structural changes persist across navigation policies using structural references, frozen small-language-model navigation policies, and trained MINERVA policies.
Across both frozen and trained policies, restoration substantially reduces released-answer accuracy and trajectory fidelity, while graph-consistent evaluation recovers only part of the loss.
These results show that benchmark graph construction is part of the navigation task specification rather than merely an implementation detail and can materially shape measured behavior even when the released QA task is held fixed.
\end{abstract}

\section{Introduction}
\label{sec:introduction}

Multi-hop knowledge-graph question answering (KGQA) asks a model to combine a natural-language question with a sequence of graph relations to reach one or more answer entities.
PathQuestion (PQ) and PathQuestion-Large (PQL), introduced with the Interpretable Reasoning Network (IRN)~\cite{zhou2018interpretable}, are long-standing benchmarks for this setting and have been used to study multi-relation reasoning over compact knowledge graphs.
For graph-navigation methods such as MINERVA~\cite{das2018go}, however, the graph is not passive background context: it directly defines the legal actions available at every reasoning step.
A benchmark can therefore become easier, harder, or simply different when its graph is filtered, even if the question text and intended relation chain do not change.
%
%
%
The present study asks a different question: \emph{what changes when the released Question-Answer (QA) tasks are placed back into the larger source/original knowledge graphs that accompany the benchmark artifacts?}
We treat this graph restoration as a controlled intervention rather than as a benchmark redesign.

Each PQ/PQL question specifies a topic entity and an intended sequence of relation types leading toward one or more answers.
We refer to this sequence as a \emph{relation chain}, and to a concrete alternating sequence of entities and relations realizing that chain in the graph as a \emph{path}.
Related surface questions that correspond to the same underlying navigation problem are grouped into a \emph{structural family} to prevent related variants from crossing split boundaries and to give each family equal weight in evaluation.
For each QA instance, we keep fixed the natural-language question, topic entity, evaluation horizon, released valid-answer set, one or more released relation chains, structural-family identity, and split assignment.
We then change only the active navigation graph from the released task-filtered graph $G_F$ to the distributed source graph $G_S$.
The primary evaluation continues to use the released answer and path annotations in both conditions, isolating the effect of changing the graph from changes to the evaluation target.
Separately, we enumerate \emph{graph-consistent} answers and paths: endpoints and path realizations obtained by following the same released relation chains in the active graph.

This distinction is important because PQ/PQL contain repeated surface variants, multiple valid answers, cyclic or immediately backtracking reference paths, and, in PQL, a small number of QA instances with more than one released relation chain.
Collapsing these cases into a single designated path, removing repeated-node trajectories, or evaluating every textual variant as an independent structural problem can distort both the measured difficulty of the benchmark and the apparent behavior of the evaluated models.
We therefore reconstruct the benchmarks using family-level split control and multi-reference annotations while preserving the released trajectories.

\textbf{Contributions}.
Our contributions are threefold:
\begin{itemize}
    \item \textsc{Intervention}.
    We formulate source restoration as a matched intervention that isolates the effect of graph construction by holding the released QA task fixed while changing only the active navigation graph.

    \item \textsc{Reconstruction}.
    We provide a family-controlled reconstruction of PQ/PQL that preserves multiple released references and cyclic trajectories while making graph-induced answer and path multiplicity explicit.
    As the original benchmarks lack an explicit redistribution license, we release the reconstruction code rather than the reconstructed datasets.\footnote{\url{https://github.com/HalcyonSolutions/PathQuestion-Family}}
    The code reproduces the benchmark versions used in this study from the original released artifacts.

    \item \textsc{Empirical Evidence}.
    We evaluate the intervention using structural references, frozen small-language-model navigation policies, and trained MINERVA policies, measuring both answer accuracy and trajectory fidelity with the path-fidelity metrics introduced for traceable KG navigation in prior work~\cite{hernandez2026theseus}.
    We complement these fixed released targets with graph-consistent diagnostics that identify newly realizable answers and paths in the restored graph.
\end{itemize}
Across these settings, source restoration consistently changes answer reachability and trajectory fidelity, showing that benchmark graph construction is a consequential part of the navigation task specification.


\section{Related Work and Benchmark History}
\label{sec:related}

\paragraph{Original construction and mixed-hop variants.}
PathQuestion and PathQuestion-Large were introduced with IRN as separate 2-hop and 3-hop benchmarks generated from Freebase subsets by extracting relation paths and rendering them through handcrafted templates with synonym and paraphrase variation~\cite{zhou2018interpretable}.
IRN characterizes PQL as the more challenging variant because it uses a larger KG while providing fewer training instances.
Thus, ``Large'' refers to the graph rather than to the number of QA examples.
The released artifacts additionally distinguish hop-specific task-filtered KGs from larger source/original KGs, a distinction that motivates the controlled restoration studied here.
UHop~\cite{chen2019uhop} later combined the 2-hop and 3-hop variants as PQ+ and PQL+ to evaluate unrestricted-hop reasoning and learned stopping, while also noting the benchmarks' small graphs, repetitive relations, and template-generated language.
SRN~\cite{qiu2020stepwise} adopted a similar mixed-hop setting under the names PQ-M and PQL-M, formulating the task as weakly supervised sequential decision making.
Our reconstruction follows this mixed-hop setting while adding family-level split control and retaining all released annotation references associated with each reconstructed QA instance.
Because these systems use different multi-hop reasoning formulations, their reported mixed-hop results should not be interpreted as measuring a common notion of explicit graph-navigation performance.
Appendix~\ref{app:historical_mixed} discusses this distinction when relating their published results to our reconstruction.

\paragraph{Multiple answers and reasoning paths.}
Answer multiplicity is already part of the original benchmark design: IRN evaluates predictions against a valid answer set for one-to-many relations~\cite{zhou2018interpretable}.
MRP-QA~\cite{wang2022mrpqa} further studies alternative reasoning paths in PQL by enumerating paths of at most three hops from the topic entity to the answer, reporting multiple valid paths for substantial portions of PQL-2H and PQL-3H, 12.5\% and 45.2\% respectively.
Our focus is therefore not the existence of multiple answers or paths, but how their graph-realizable multiplicity changes when the released task graph is replaced by the larger source graph while the released relation-chain annotations remain fixed.

\paragraph{Incomplete-KG and robustness studies.}
Prior work has also changed the graph supporting PQ/PQL, but mainly by deleting information from the released task graph.
IRN evaluates an incomplete-KG condition after removing half of the PQ-2H triples~\cite{zhou2018interpretable}, and QA2MN~\cite{li2022questionaware} similarly evaluates robustness when only half of the original task-graph triples are retained.
These interventions ask whether a model can tolerate missing support.
They are not equivalent to source restoration, which moves in the opposite direction and can introduce additional branches, intermediate entities, answer endpoints, and shortcuts that were suppressed by benchmark filtering.

\paragraph{Later PQ/PQL use and structural analysis.}
Later work continues to use the hop-specific and mixed variants.
For example, HyperMR reports PQ-2H, PQ-3H, PQ-M and their PQL counterparts~\cite{wang2024hypermr}.
Prior work on traceable multi-hop graph navigation~\cite{hernandez2026theseus} established the structural calibration tools reused here: exact RW-Ans$_p$, its Monte Carlo estimate RW-Ans$_{\mathrm{MC}}$, the answer-oracle Shortest Path reference, and the PED, RED, F1$_{\mathrm{SG}}$, and F1$_{\mathrm{REL}}$ trajectory metrics.
That study's Appendix~C already applies the random-walk diagnostics and local-navigation statistics to release-derived mixed PQ/PQL graphs.
A concurrent study of small language models evaluates isolated iterative graph navigation beyond answer accuracy~\cite{hernandez2026path}.
We reuse both the established calibration suite and the frozen-SLM navigation protocol, but apply them to a different experimental question: how navigation changes under a matched filtered-versus-source graph intervention with fixed released QA semantics.

\paragraph{Novelty boundary.}
Prior work therefore covers PQ/PQL construction, mixed-hop stopping, templated language, incomplete-KG robustness, answer multiplicity, alternative reasoning paths, and structural navigation diagnostics.
To the best of our review, however, we did not find a study that holds the released questions, topic entities, relation-chain annotations, split/family assignment, and released answer target fixed while replacing the released task-filtered graph with the distributed source/original knowledge graph.
Our contribution is this matched environmental substitution and its effect on benchmark navigability, rather than a first general audit of PQ/PQL.

\section{Problem Formulation}
\label{sec:preliminaries}

\paragraph{Question-conditioned graph navigation.}
Let $G=(\mathcal{E},\mathcal{R},\mathcal{T})$ be a directed knowledge graph with entities $\mathcal{E}$, relations $\mathcal{R}$, and labeled triples $\mathcal{T}\subseteq\mathcal{E}\times\mathcal{R}\times\mathcal{E}$.
A QA instance $q$ contains a natural-language question, a topic entity $e_s$, a released valid-answer set $A_{\mathrm{released}}(q)$, and one or more released relation chains $\mathcal{R}^{\star}(q)$.
A navigation policy starts at $e_s$ and selects among the labeled outgoing actions exposed by the active graph until the fixed evaluation horizon is reached.

\paragraph{Filtered and source graphs.}
We denote the released task-filtered graph by $G_F$ and the larger distributed source/original knowledge graph by $G_S$.
Our primary intervention keeps $q$, $e_s$, $A_{\mathrm{released}}(q)$, and $\mathcal{R}^{\star}(q)$ fixed while replacing $G_F$ with $G_S$.
Thus the primary comparison is
\begin{equation}
(G_F,A_{\mathrm{released}})\quad\longleftrightarrow\quad(G_S,A_{\mathrm{released}}),
\label{eq:intervention}
\end{equation}
with the same family-level train/dev/test assignment in both conditions.
We deliberately avoid calling $G_S$ ``full Freebase,'' because the released source files are benchmark artifacts rather than a claim to contain the complete original knowledge graph.

\paragraph{Graph-consistent answer and path sets.}
For an active graph $G$, let $A_{\mathrm{graph}}(q)$ be the set of endpoints obtained by starting at $e_s$ and following any released chain in $\mathcal{R}^{\star}(q)$ under the benchmark's unrestricted-walk semantics.
Likewise, let $\mathcal{P}_{\mathrm{graph}}(q)$ contain every entity-level realization of those chains in $G$.
The released answer set $A_{\mathrm{released}}(q)$ remains the primary benchmark target, while $A_{\mathrm{graph}}(q)$ and $\mathcal{P}_{\mathrm{graph}}(q)$ are secondary structural quantities used to measure graph-induced answer and path expansion.
This separation prevents source restoration from automatically promoting every newly reachable endpoint to natural-language semantic gold.

\paragraph{Adopted navigation diagnostics.}
We adopt the structural and trajectory-evaluation protocol of prior work~\cite{hernandez2026theseus} rather than introduce a new metric suite.
RW-Ans$_p$ is the exact fixed-horizon probability that a uniform random walker over labeled legal actions terminates in the valid answer set, while RW-Ans$_{\mathrm{MC}}$ is its Monte Carlo estimate based on sampled walks.
The same work defines an answer-oracle Shortest Path structural reference that is given the topic entity, navigation graph, and valid answer set, but not the annotated reasoning path, and deterministically selects a shortest route to any valid answer.
We likewise reuse Path Edit Distance (PED) and entity/subgraph overlap F1$_{\mathrm{SG}}$ for path fidelity: PED is the Levenshtein distance between the ordered predicted and reference edge sequences, while $F1_{\mathrm{SG}}$ measures their order-invariant edge overlap.
Together, these metrics capture complementary aspects of trajectory fidelity: ordered path agreement and order-invariant edge overlap.
When multiple released entity-level paths or relation chains are valid, our multi-annotated evaluation takes the minimum edit distance or maximum overlap over the valid released references.
This audit therefore changes the graph condition and valid-reference bookkeeping, not the definitions of RW-Ans$_p$, RW-Ans$_{\mathrm{MC}}$, the Shortest Path reference, or the path-fidelity metrics.
As previously noted~\cite{hernandez2026theseus}, the Shortest Path reference is not a path-fidelity upper bound because a graph-theoretic shortcut can reach a correct answer while disagreeing with the released reasoning chain.

\paragraph{Family-aware aggregation.}
Generated surface variants are grouped into structural question families for split control and aggregation.
If family $f$ contains $|f|$ evaluation instances with per-instance metric $m_i$, the primary family-macro statistic is
\begin{equation}
\overline m_{\mathrm{fam}}
=\frac{1}{|\mathcal{F}|}\sum_{f\in\mathcal{F}}
\frac{1}{|f|}\sum_{i\in f}m_i.
\label{eq:familymacro}
\end{equation}
Family-macro values are primary for the structural comparisons in this paper so that large surface-form families do not dominate the result.

\section{Benchmark Reconstruction and Graph Intervention}
\label{sec:dataset}

\paragraph{QA reconstruction.}
Table~\ref{tab:reconstruction}, Panel A summarizes the released-record counts, reconstructed QA-instance counts, family counts, and hop composition used by the audit.
PQ contains one repeated exact QA group whose released answers are unioned, while PQL contains eight same-question/topic/hop cases with two released relation chains, with both chains and their associated paths retained as valid references.
Structural families are assigned to train/dev/test using an 80/10/10 split, preventing structurally shared variants from crossing split boundaries.
Because splitting is performed at the family level, the corresponding QA-instance proportions can differ slightly from this ratio.

\begin{table}[t]
    \caption{
    Reconstructed QA statistics and source-restoration effects under fixed released relation-chain semantics.
    Bracketed values in Panel A denote QA instances.
    Panel B reports the number of affected structural families or QA instances relative to the corresponding totals in Panel A.
    }
    \label{tab:reconstruction}
    \centering
    \small
    \setlength{\tabcolsep}{5pt}
    \begin{tabular}{lrr}
        \toprule
        \textbf{Statistic} & \textbf{PQ} & \textbf{PQL} \\
        \midrule
        \multicolumn{3}{l}{\textsc{A. Reconstructed QA Statistics}} \\
        Released QA records             & 7,106 & 2,625 \\
        Reconstructed QA instances      & 7,105 & 2,114 \\
        Structural QA families          & 2,923 & 1,530 \\
        \multicolumn{3}{l}{\textit{A.1 - QA Hop Structure}} \\
        2-hop families                  & 611 [1,908] & 584 [1,168] \\
        3-hop families                  & 2,312 [5,197] & 946 [946] \\
        \multicolumn{3}{l}{\textit{A.2 - QA Train Split}} \\
        Train families                  & 2,337 [5,674] & 1,223 [1,690] \\
        Dev families                    & 292 [707] & 153 [211] \\
        Test families                   & 294 [724] & 154 [213] \\
        
        \midrule
        \multicolumn{3}{l}{\textsc{B. Expansion under Source Restoration}} \\
        \multicolumn{3}{l}{\textit{B.1 - New graph-consistent answers}} \\
        Structural QA families          & 2,158 / 2,923 & 633 / 1,530 \\
        QA instances                    & 5,325 / 7,105 & 866 / 2,114 \\
        Added answer assignments        & 13,307 & 13,541 \\
        \multicolumn{3}{l}{\textit{B.2 - New graph-consistent paths}} \\
        Structural QA families          & 2,437 / 2,923 & 647 / 1,530 \\
        QA instances                    & 5,982 / 7,105 & 883 / 2,114 \\
        Added path realizations         & 50,230 & 14,147 \\
        \bottomrule
    \end{tabular}
\end{table}

\paragraph{Graph conditions.}
Table~\ref{tab:graph_restoration} reports global filtered-versus-source graph statistics, including scale, connectivity, degree distribution, outgoing relation diversity, and dead-end prevalence.
For PQ, source restoration preserves the 13-relation vocabulary while substantially increasing graph size, degree, and connectivity, with the largest weakly connected component (WCC) growing from 93.75\% to the entire graph.
PQL changes more substantially: the source graph expands both the entity population and relation vocabulary, develops a much heavier out-degree tail, and increases the largest-WCC coverage from 38.69\% to 98.33\% despite containing more disconnected components overall.
These global properties characterize the navigation environments but do not determine question-conditioned difficulty.
Section~\ref{sec:experiments} therefore measures the action spaces encountered along benchmark-relevant states.

\begin{table}[t]
    \caption{
    Global graph statistics for the released filtered graphs and distributed source/original KGs.
    }
    \label{tab:graph_restoration}
    \centering
    \small
    \setlength{\tabcolsep}{2pt}
    \begin{tabular}{lrrrr}
        \toprule
        \multirow{2}{*}{\textbf{Graph property}}
        & \multicolumn{2}{c}{\textbf{PQ}}
        & \multicolumn{2}{c}{\textbf{PQL}} \\
        \cmidrule(lr){2-3}
        \cmidrule(lr){4-5}
        & \multicolumn{1}{c}{\textbf{$G_F$}}
        & \multicolumn{1}{c}{\textbf{$G_S$}}
        & \multicolumn{1}{c}{\textbf{$G_F$}}
        & \multicolumn{1}{c}{\textbf{$G_S$}}\\
        \midrule
        Entities $|\mathcal{E}|$        & 2,256   & 75,043  & 6,505   & 416,249 \\
        Relations $|\mathcal{R}|$       & 13      & 13      & 411     & 3,076 \\
        Triples $|\mathcal{T}|$         & 3,377   & 375,444 & 5,597   & 831,940 \\
        WCCs                            & 46      & 1       & 1,346   & 3,229 \\
        Largest WCC (\%)                & 93.75   & 100.00  & 38.69   & 98.33 \\
        Avg. out-degree                 & 1.50    & 5.00    & 0.86    & 2.00 \\
        Out-degree p90/p95/p99          & 3/4/6   & 9/10/14 & 1/2/4   & 3/4/12 \\
        Max. out-degree                 & 10      & 37      & 115     & 26,023 \\
        Avg. distinct out-rel.          & 1.40    & 4.08    & 0.66    & 1.12 \\
        Zero out-degree (\%)            & 27.48   & 10.19   & 35.73   & 23.14 \\
        Clustering coefficient          & 0.0633  & 0.0278  & 0.0006  & 0.0831 \\
        \bottomrule
    \end{tabular}
\end{table}

\paragraph{Cycles and backtracking.}
Table~\ref{tab:released_path_cycles} summarizes entity revisits and immediate backtracking in the original released annotations, before question-family aggregation.
An entity revisit occurs when a path visits an entity more than once, whereas an immediate backtrack returns to the entity visited two positions earlier.
Both behaviors are prevalent, particularly in PQL three-hop annotations, where nearly all reference paths revisit an entity.
Relatedly, the topic entity itself belongs to the released answer set for 33.0\% of PQL released records, so answer reachability can include a valid zero-length solution even when the released reasoning annotation follows a multi-hop walk.
Appendix~\ref{sec:appendix_topic_answer} provides the full breakdown.
These behaviors expose a benchmark-design concern for future multi-hop KGQA datasets because they can weaken the alignment between the surface question, the annotated reasoning chain, and the navigation actually required to reach a valid answer.
In particular, cyclic or self-returning annotations can make the nominal multi-hop trajectory unnecessary for answer retrieval, raising questions about whether the assigned hop count reflects required multi-hop reasoning.
This concern is especially apparent when the topic entity itself is a released answer, since the benchmark can admit a valid zero-length solution despite providing a multi-hop reference trajectory.
Appendix~\ref{sec:appendix_wording} separately examines wording-related irregularities that can further affect the interpretation of the released questions.
For the present audit, however, these properties are part of the released benchmark and must remain unchanged.
We therefore preserve the released walk semantics rather than imposing a no-repeated-entity constraint, which would alter the reference reasoning trajectories and the task being audited.

\begin{table}[t]
\caption{
    Entity revisits and immediate backtracking in the released PQ/PQL reference paths.
}
\label{tab:released_path_cycles}
\centering
\small
\setlength{\tabcolsep}{5pt}
\begin{tabular}{llrrr}
\toprule
Dataset & Hops & Paths & Revisited & Backtracked \\
\midrule
PQ  & 2   & 1,908 & 120 (6.3\%)    & 117 (6.1\%) \\
    & 3   & 5,198 & 2,488 (47.9\%) & 2,416 (46.5\%) \\
    & All & 7,106 & 2,608 (36.7\%) & 2,533 (35.6\%) \\
\midrule
PQL & 2   & 1,594 & 822 (51.6\%)   & 420 (26.3\%) \\
    & 3   & 1,031 & 982 (95.2\%)   & 858 (83.2\%) \\
    & All & 2,625 & 1,804 (68.7\%) & 1,278 (48.7\%) \\
\bottomrule
\end{tabular}
\end{table}

\paragraph{Answer and path restoration.}
With the released reasoning relation-chain annotations preserved, Table~\ref{tab:reconstruction}, Panel B shows that restoring the source graph while holding the released relation chains fixed expands the set of graph-consistent endpoints and path realizations for a large fraction of both benchmarks.
The expansion is especially pronounced for PQ path realizations and PQL answer multiplicity, showing that endpoint multiplicity and trajectory multiplicity are distinct structural effects.
Appendix~\ref{sec:appendix_graph_expansion} illustrates this distinction with concrete examples in which the same released relation chain reaches additional graph-consistent answers after source restoration.
No released answer or path is lost under source restoration in the reconstructed audit data, so the intervention strictly expands the available graph-consistent answers and path realizations.

\section{Results}
\label{sec:experiments}

\begin{table*}[t]
    \caption{
Primary family-macro navigation results on PathQuestion (PQ). 
Structural references contextualize unguided answer reachability and shortest-path fidelity; 
frozen SLMs are evaluated without task-specific training, and MINERVA is trained on the active KG. 
The two Source MINERVA rows distinguish the matched filtered-dev configuration from a source-dev-tuned control.
}
    \label{tab:pq_policy_results}
    \centering
    \small
    \begin{tabular}{lllllll}
        \toprule
        \multirow{2}{*}{\textbf{Method}} 
        & \multirow{2}{*}{\textbf{Active KG}} 
        & \multirow{2}{*}{\textbf{Tuning}}
        & \multicolumn{1}{c}{\textbf{$A_{\mathrm{released}}$}}
        & \multicolumn{2}{c}{\textbf{$P_{\mathrm{released}}$}}
        & \multicolumn{1}{c}{\textbf{$A_{\mathrm{graph}}$}} \\
        \cmidrule(lr){4-4}
        \cmidrule(lr){5-6}
        \cmidrule(lr){7-7}
        & &
        & \textbf{H@1}$\uparrow$ 
        & \textbf{PED}$\downarrow$ 
        & \textbf{F1$_{\mathrm{SG}}$}$\uparrow$ 
        & \textbf{H@1}$\uparrow$ \\
        
        \midrule
        \multicolumn{6}{l}{\textit{Structural References}} \\
        \multirow{2}{*}{$\textsc{RW-Ans}_{\mathrm{MC}}$}
        & Filtered & -- & \numstats{0.175420,0.175748,0.173940} & \numstats{2.057279,2.050057,2.055179} & \numstats{0.340163,0.342258,0.340701} & \numstats{0.175709,0.175998,0.174252} \\
        & Source & -- & \numstats{0.068112,0.067630,0.068109} & \numstats{2.624348,2.622775,2.626086} & \numstats{0.082853,0.083324,0.081552} & \numstats{0.120241,0.118917,0.120607}\\
        \multirow{2}{*}{Shortest Path Oracle} 
        & Filtered & -- & \num{1.000} & \num{1.068027} & \num{0.720181} & \num{1.000}\\
        & Source & -- & \num{1.000} & \num{1.772109} & \num{0.464512} & \num{1.000} \\

        \midrule
        \multicolumn{6}{l}{\textit{Frozen policies}} \\
        \multirow{2}{*}{Qwen3 (8B)}
        & Filtered & -- & \num{0.638888888888889} & \num{0.7797619047619048} & \num{0.7707577475434618} & \num{0.6422902494331066} \\
        & Source   & -- & \num{0.3137755102040816} & \num{1.9382086167800454} & \num{0.36848072562358275} & \num{0.4781746031746032} \\
        \addlinespace
        \multirow{2}{*}{Ministral 3 (8B)}
        & Filtered & -- & \num{0.5802154195011339} & \num{0.9571995464852608} & \num{0.7432917611489039} & \num{0.5830498866213152} \\
        & Source   & -- & \num{0.30328798185941047} & \num{1.8239795918367347} & \num{0.41457860922146633} & \num{0.4869614512471655} \\
        \addlinespace
        \multirow{2}{*}{Gemma 4 (E4B)}
        & Filtered & -- & \num{0.5898526077097506} & \num{0.8202947845804989} & \num{0.76921768707483} & \num{0.5915532879818594} \\
        & Source   & -- & \num{0.28883219954648526} & \num{1.8979591836734695} & \num{0.3761526832955404} & \num{0.47335600907029474} \\
        \midrule
        \multicolumn{6}{l}{\textit{Trainable policy}} \\
        \multirow{3}{*}{MINERVA}
        & Filtered & Filtered dev & \numstats{0.8450,0.8209,0.8393} & \numstats{0.4524,0.5320,0.5748} & \numstats{0.8617,0.8382,0.8228} & \numstats{0.8467,0.8248,0.8421}\\
        & Source   & Filtered dev & \numstats{0.5388,0.5439,0.5626} & \numstats{1.9651,1.9623,1.9328} & \numstats{0.3640,0.3664,0.3849} & \numstats{0.7080,0.6902,0.7058} \\
        & Source   & Source dev   & \numstats{0.5592,0.5456,0.5760} & \numstats{1.8827,1.8886,1.8716} & \numstats{0.3906,0.3898,0.3980} & \numstats{0.6862,0.7120,0.7211} \\
        \bottomrule
    \end{tabular}
\end{table*}

\begin{table*}[t]
    \caption{
Primary family-macro navigation results on PathQuestion-Large (PQL). 
Structural references contextualize unguided answer reachability and shortest-path fidelity; 
frozen SLMs are evaluated without task-specific training, and MINERVA is trained on the active KG. 
The two Source MINERVA rows distinguish the matched filtered-dev configuration from a source-dev-tuned control.
}
    \label{tab:pql_policy_results}
    \centering
    \small
    \begin{tabular}{lllllll}
        \toprule
        \multirow{2}{*}{\textbf{Method}} 
        & \multirow{2}{*}{\textbf{Active KG}} 
        & \multirow{2}{*}{\textbf{Tuning}}
        & \multicolumn{1}{c}{\textbf{$A_{\mathrm{released}}$}}
        & \multicolumn{2}{c}{\textbf{$P_{\mathrm{released}}$}}
        & \multicolumn{1}{c}{\textbf{$A_{\mathrm{graph}}$}} \\
        \cmidrule(lr){4-4}
        \cmidrule(lr){5-6}
        \cmidrule(lr){7-7}
        & &
        & \textbf{H@1}$\uparrow$ 
        & \textbf{PED}$\downarrow$ 
        & \textbf{F1$_{\mathrm{SG}}$}$\uparrow$ 
        & \textbf{H@1}$\uparrow$ \\

        \midrule
        \multicolumn{6}{l}{\textit{Structural References}} \\
        \multirow{2}{*}{$\textsc{RW-Ans}_{\mathrm{MC}}$}
        & Filtered & -- & \numstats{0.540714,0.542143,0.539058} & \numstats{1.627760,1.627175,1.632208} & \numstats{0.550644,0.549254,0.548846} & \numstats{0.540714,0.542143,0.539058} \\
        & Source & -- & \numstats{0.076883,0.078377,0.077662} & \numstats{2.482792,2.478669,2.485487} & \numstats{0.131537,0.132844,0.131449} & \numstats{0.110649,0.113312,0.111136}\\
        \multirow{2}{*}{Shortest Path Oracle} 
        & Filtered & -- & \num{1.000} & \num{1.681818} & \num{0.484848} & \num{1.000}\\
        & Source & -- & \num{1.000} & \num{1.779221} & \num{0.445887} & \num{1.000} \\

        \midrule
        \multicolumn{6}{l}{\textit{Frozen policies}} \\
        \multirow{2}{*}{Qwen3 (8B)}
        & Filtered & -- & \num{0.8116883116883117} & \num{1.0227272727272727} & \num{0.7653679653679654} & \num{0.8116883116883117} \\
        & Source   & -- & \num{0.3051948051948052} & \num{2.1103896103896105} & \num{0.30454545454545456} & \num{0.474025974025974} \\
        \addlinespace
        \multirow{2}{*}{Ministral 3 (8B)}
        & Filtered & -- & \num{0.7662337662337663} & \num{1.2694805194805194} & \num{0.7309523809523809} & \num{0.7662337662337663} \\
        & Source   & -- & \num{0.4090909090909091} & \num{1.87987012987013} & \num{0.4027056277056277} & \num{0.6363636363636364} \\
        \addlinespace
        \multirow{2}{*}{Gemma 4 (E4B)}
        & Filtered & -- & \num{0.7694805194805194} & \num{0.948051948051948} & \num{0.749025974025974} & \num{0.7694805194805194} \\
        & Source   & -- & \num{0.38961038961038963} & \num{1.853896103896104} & \num{0.4113636363636364} & \num{0.5616883116883117} \\
        \midrule
        \multicolumn{6}{l}{\textit{Trainable policy}} \\
        \multirow{3}{*}{MINERVA}
        & Filtered & Filtered dev & \numstats{0.9416,0.9448,0.9351} & \numstats{1.2857,1.2662,1.2955} & \numstats{0.7353,0.7255,0.7179} & \numstats{0.9416,0.9448,0.9351} \\
        & Source   & Filtered dev & \numstats{0.8084,0.8279,0.8539} & \numstats{2.0552,2.0617,2.0292} & \numstats{0.3367,0.3209,0.3325} & \numstats{0.8182,0.8409,0.8604} \\
        & Source   & Source dev   & \numstats{0.8377,0.8149,0.8539} & \numstats{1.8831,2.0357,1.9448} & \numstats{0.4419,0.3536,0.4119} & \numstats{0.8636,0.8506,0.8669} \\
        \bottomrule
    \end{tabular}
\end{table*}

\paragraph{Protocol.}
We evaluate the mixed-hop test splits under both the filtered and restored source graph conditions.
Family-macro aggregation follows Eq.~\eqref{eq:familymacro} and is used as the primary aggregation throughout the main evaluation.
For multi-answer questions, an endpoint is counted as correct if it belongs to the corresponding valid answer set.
All primary evaluations use directed outgoing actions with a self-loop, a fixed three-action horizon, and the mixed-hop evaluation setting.
For all model-based comparisons, the entity and relation ID vocabularies are fixed to those defined by $G_S$, which contain all entities and relations present in $G_F$.
Changing from $G_F$ to $G_S$ therefore changes the available triples and outgoing actions, not the representable entity or relation vocabulary.
Following prior work~\cite{hernandez2026theseus}, RW-Ans$_p$ uses uniform labeled actions and RW-Ans$_{\mathrm{MC}}$ uses 100 sampled walks per question for each of seeds 0, 42, and 100.
The answer-oracle Shortest Path reference receives the released answer set, ignores self-loop actions during breadth-first search, and is evaluated against the released path references.
For the frozen policies, we evaluate Gemma 4 E4B~\cite{team2026gemma}, Ministral 3 8B~\cite{liu2026ministral}, and Qwen3 8B~\cite{yang2025qwen3}.
Each model is used as a frozen policy through the same isolated iterative-navigation interface in both graph conditions, without task-specific parameter updates~\cite{hernandez2026path}.
MINERVA is trained on the active KG with seeds 0, 42, and 100.
\footnote{The THESEUS project (\url{https://github.com/HalcyonSolutions/THESEUS}) provides links to the navigation implementations, including the frozen SLM policies used in \citet{hernandez2026path}, and the trained MINERVA checkpoints for PQ/PQL.}
For the matched filtered-versus-source comparison, we use the configuration selected on the filtered development set in both graph conditions so that the active graph is the only changed experimental condition.
Because this configuration may favor the filtered environment, we additionally select a configuration on the source development set using the same hyperparameter search space and tuning budget.
This source-dev-selected control measures how much of the source-graph degradation can be mitigated by condition-matched hyperparameter selection.
Released answers $A_{\mathrm{released}}$ and paths $P_{\mathrm{released}}$ remain the primary fixed targets, while $A_{\mathrm{graph}}$ and $P_{\mathrm{graph}}$ provide complementary graph-relative diagnostics.

%


\subsection{Structural references separate answer reachability from path fidelity}

Tables~\ref{tab:pq_policy_results} and~\ref{tab:pql_policy_results} show that source restoration sharply reduces both released-answer reachability and trajectory fidelity for RW-Ans$_{\mathrm{MC}}$ on PQ and PQL.
The contrast is especially pronounced for PQL, where RW-Ans$_{\mathrm{MC}}$ attains $54.1\%$ released-answer H@1 on the filtered graph despite using unguided uniform navigation, before falling to $7.8\%$ under source restoration.
Its path fidelity deteriorates in parallel, indicating that the filtered PQL graph makes both released endpoints and the released reasoning trajectories unusually accessible to unguided navigation.
Under source restoration, random-walk trajectory fidelity also moves close to the low-fidelity end of both metrics, with PED approaching $3$ and F1$_{\mathrm{SG}}$ approaching $0$ under the fixed three-action horizon.
This deterioration is more pronounced for PQ than PQL, indicating particularly weak agreement between unguided source-graph trajectories and the released PQ reasoning paths.
The corresponding exact random-walk calibration and additional structural statistics are reported in Appendix~\ref{sec:appendix_structural_calibration}.

The answer-oracle Shortest Path reference reaches a released answer by construction, so its H@1 remains $1.00$ in both graph conditions.
Its trajectory-fidelity metrics are therefore the more informative comparison.
For PQ, source restoration substantially degrades Shortest Path agreement with the released trajectories.
PQL exhibits a different pattern.
On the filtered graph, Shortest Path has lower path fidelity than RW-Ans$_{\mathrm{MC}}$, despite its guaranteed endpoint success.
Notably, an answer-informed shortest route therefore agrees less closely with the released reasoning trajectories than an unguided uniform random walk.
This result is consistent with the cyclic and backtracking structure documented in Section~\ref{sec:dataset}, where the shortest route to a released answer need not reproduce the annotated multi-hop walk.
After source restoration, Shortest Path fidelity degrades only slightly, while random-walk fidelity deteriorates sharply, leaving the shortest routes substantially more faithful to the released trajectories.
This reversal further shows that endpoint accessibility and agreement with the released reasoning trajectory are distinct properties whose relationship depends on the active graph.
Complete Shortest Path trajectory metrics are reported in Appendix~\ref{sec:appendix_oracle_fidelity}.

\subsection{Frozen and learned policies remain sensitive to the graph condition}

Tables~\ref{tab:pq_policy_results} and~\ref{tab:pql_policy_results} additionally report results for both frozen and trainable policies.
Across all three frozen SLMs, source restoration reduces released-answer accuracy and trajectory fidelity on both benchmarks.
The agreement in direction across independently frozen policies indicates that the effect is not specific to one SLM.

MINERVA shows the same qualitative sensitivity after training on each active graph.
The PQL result is particularly informative because source-graph endpoint accuracy remains comparatively strong while agreement with the released reasoning trajectories deteriorates substantially.
Selecting hyperparameters on the source development graph improves some source-graph metrics relative to the matched filtered-dev configuration but does not recover filtered-graph performance.
The filtered-to-source change therefore persists after allowing the trainable policy a condition-matched tuning control.

Across the filtered conditions, PQL also yields higher released-answer H@1 than PQ for RW-Ans$_{\mathrm{MC}}$, all three frozen SLMs, and MINERVA despite its larger graph and broader relation vocabulary.
This consistent reversal shows that \textbf{overall graph size} and relation-vocabulary size are not reliable proxies for \textbf{released-answer accessibility}, which is shaped by the structure of the active navigation graph.

\subsection{Graph-consistent evaluation explains only part of the degradation}

Graph-consistent evaluation distinguishes disagreement with the released answer and path references from endpoints and trajectories that remain consistent with the released relation chains but become newly realizable after source restoration.
Across random walking, the frozen policies, and MINERVA, evaluating against $A_{\mathrm{graph}}$ recovers part of the released-answer loss on the source graph.
This recovery shows that some predictions counted as incorrect under $A_{\mathrm{released}}$ nevertheless reach endpoints consistent with the same released relation chains in the restored graph.
The effect is not uniform across models or benchmarks.
For MINERVA, the recovery is substantial on PQ but comparatively small on PQL, indicating that graph-induced answer expansion explains only part of the filtered-to-source performance gap.

The same distinction applies to trajectories.
Evaluation against $P_{\mathrm{graph}}$ recovers part of the source-graph path-fidelity loss, particularly on PQ, while the recovery for MINERVA on PQL is much smaller.
Appendix~\ref{sec:appendix_graph_path_eval} reports the complete graph-consistent trajectory evaluation.
We therefore treat $A_{\mathrm{graph}}$ and $P_{\mathrm{graph}}$ as complementary graph-relative diagnostics rather than replacements for the fixed released benchmark targets.

\subsection{Cross-graph transfer reveals environment specialization}

Appendix~\ref{sec:appendix_cross_graph_transfer} evaluates trained MINERVA checkpoints across graph conditions without destination-graph retuning.
Policies trained on $G_F$ degrade substantially when transferred to $G_S$, whereas policies trained on $G_S$ transfer more favorably to $G_F$.
Because the released targets remain available and the entity and relation vocabularies are fixed across conditions, this asymmetry cannot be explained by target disappearance or an out-of-vocabulary mismatch.
The transfer results therefore provide additional evidence that learned navigation policies specialize to the graph environment in which they are trained.

\paragraph{Interpretation.}
The combined structural, frozen-policy, trained-policy, graph-consistent, and transfer results do not imply that a larger graph is uniformly harder to navigate.
Instead, they show that changing the active graph changes the decision problem induced by the same released QA task.
Branching, reachable-state volume, answer and path multiplicity, unguided answer reachability, learned endpoint accuracy, and trajectory fidelity all respond to this intervention.
The active graph should therefore be treated as part of the benchmark specification rather than as an interchangeable implementation artifact.

\paragraph{Community implications.}
The released PQ/PQL graphs remain useful controlled environments, but results obtained on them should be interpreted in light of their graph construction, generated language, answer annotations, and path conventions.
In particular, high accuracy on the filtered graph does not establish equivalent navigation behavior on the distributed source KG or faithful recovery of the released reasoning trajectories.
Appendix~\ref{app:implications} summarizes cautions for interpreting and comparing results obtained on PQ/PQL.

\section{Conclusion}
\label{sec:conclusion}

We audited PathQuestion and PathQuestion-Large by treating source-graph restoration as a controlled intervention that changes the active navigation environment while keeping the released QA task fixed.
Restoration substantially expands the effective navigation search space, increases graph-consistent answer and path multiplicity, reduces unguided reachability to released answers, and lowers shortest-path agreement with the released reasoning traces.
These structural changes persist across three frozen SLM policies and trained MINERVA policies, showing that the effect is not limited to a model-free calibration.

The completed learned-policy experiments sharpen the distinction between endpoint success and trajectory fidelity.
MINERVA retains relatively strong source-graph answer accuracy on PQL while its agreement with the released paths degrades substantially.
Graph-consistent evaluation recovers part of the apparent answer and path loss by crediting newly realizable endpoints and trajectories that satisfy the same released relation chains, but the recovery is incomplete and strongly benchmark-dependent.
Cross-graph transfer provides an additional control.
Filtered-trained policies degrade sharply when moved to the restored source graph, whereas source-trained policies remain comparable or improve when the graph is contracted to the filtered environment.
Because the released targets remain available in both KGs, this asymmetry isolates sensitivity to the surrounding graph structure rather than target disappearance.

These results show that benchmark graph construction is a consequential part of the navigation task specification.
PQ/PQL remain useful controlled benchmarks, but their filtered graphs, cyclic path conventions, multiple valid references, and family structure must be made explicit when results are interpreted or compared.
The central implication is not that source graphs are universally harder.
It is that the same nominal multi-hop QA task can induce materially different navigation behavior when the graph defining the legal actions changes.
Future KGQA benchmarks and evaluations should therefore report graph provenance and filtering, represent multiple released and graph-consistent answers and paths explicitly, and separate released-target performance from graph-relative structural validity.

\section*{Limitations}

This study focuses on PathQuestion (PQ) and PathQuestion-Large (PQL), and its findings may not generalize directly to other multi-hop KGQA benchmarks with different graph construction
procedures, question distributions, or annotation conventions.

Our source-graph intervention uses the larger knowledge graphs distributed with the original benchmark artifacts. These graphs should not be interpreted as complete versions of Freebase.
Consequently, the analysis measures the effects of restoring the available source graphs rather than those of removing all possible benchmark filtering.

Graph-consistent answers and paths are defined by the released relation-chain annotations. 
Although this provides a reproducible structural criterion, it does not guarantee that every newly reachable endpoint is semantically correct for the natural-language question. 
We therefore retain the released answers and paths as the primary evaluation targets.

Finally, our policy-based evaluation covers a limited selection of frozen small language models and trained MINERVA policies under a fixed navigation horizon. 
Other architectures, training procedures, or adaptive stopping strategies may respond differently to graph restoration.

\section*{Ethical Considerations}

This study uses publicly released KGQA benchmarks and does not collect new personal data.
The underlying knowledge graphs may contain factual errors, omissions, or biases inherited from their sources.
We distinguish graph-consistent answers from released benchmark answers to avoid overstating model correctness or semantic validity.

\section*{Acknowledgment}
This work is partially funded by the NSTC grant number 113-2923-E-A49-001 and by MARC, the MediaTek Advanced Research Center with grant number 114A540531.

{
\small
\bibliography{reference_nonanonymous}
}
\clearpage
\newpage
\appendix

\section{Additional Benchmark Details}
\label{sec:appendix_reconstruction}

\subsection{Evaluation-instance and family semantics}

Structural families are used for split control and macro aggregation, but they are not collapsed into a single canonical natural-language training example.
The primary family-macro statistics assign each instance inverse-family-size weight so that every complete family contributes total weight one.
The corresponding evaluation-instance micro results preserve the same qualitative filtered-to-source trends and are retained in the machine-readable outputs for compatibility with the released question distribution.

\subsection{Topic--answer identity and cyclic structure}
\label{sec:appendix_topic_answer}

Table~\ref{tab:topic_answer_identity} reports how often the topic entity is itself a released answer, broken down by annotated hop count.
The phenomenon is relatively limited in PQ but is a substantial property of PQL, particularly among its 3-hop instances.
It also persists after family aggregation, affecting 76 PQ families and 592 PQL families.
For answer-based evaluation, the relevant case is whether the topic belongs to the released answer set, since any member of that set constitutes a valid endpoint.
When this occurs, an answer-oracle shortest-path procedure can legitimately select the topic itself and return a zero-length trajectory even though the released annotation follows a 2- or 3-hop walk.
For example, PQ Question~18 asks \emph{``who is the child of \texttt{shah\_shuja}'s parent?''}, and its released reference path returns to the topic entity after 2 hops.
Likewise, PQL Question~1168 asks \emph{``what is the recording of recording of \texttt{Robin\_Hood}'s recording?''}, and its released 3-hop reference path remains at the topic entity throughout.
These examples illustrate that shortest answer reachability and agreement with the annotated reasoning trajectory are distinct properties, especially in the presence of cyclic and immediate-backtracking structure.

\begin{table}[t]
\caption{Topic--answer identity in the released PQ/PQL annotations.}
\label{tab:topic_answer_identity}
\centering
\small
\setlength{\tabcolsep}{4pt}
\begin{tabular}{llrr}
\toprule
Dataset & Hops & $e_s = A_{\mathrm{released}}$ & $e_s \in A_{\mathrm{released}}$ \\
\midrule
PQ  & 2   & 117 (6.1\%) & 120 (6.3\%) \\
    & 3   & 74 (1.4\%)  & 78 (1.5\%) \\
    & All & 191 (2.7\%) & 198 (2.8\%) \\
\midrule
PQL & 2   & 420 (26.3\%) & 474 (29.7\%) \\
    & 3   & 382 (37.1\%) & 392 (38.0\%) \\
    & All & 802 (30.6\%) & 866 (33.0\%) \\
\bottomrule
\end{tabular}

\end{table}

\subsection{Graph-consistent answer and path expansion}
\label{sec:appendix_graph_expansion}

The reconstructed representation stores released answer and path references separately from graph-relative realizations of the released relation chains.
This distinction is necessary for the controlled intervention because the released answers and reference paths remain fixed across $G_F$ and $G_S$, while the set of graph-consistent endpoints and trajectories induced by those same relation chains can expand with the active graph.
Source restoration therefore does not replace the benchmark target.
Instead, it reveals additional entities and paths that satisfy the released structural annotation in the restored KG.

PQ Question~18 provides a concrete example.
The question asks \emph{``who is the child of \texttt{shah\_shuja}'s parent?''}, corresponding to a 2-hop parent-to-child relation chain.
In the filtered graph, the released realization returns to \texttt{shah\_shuja}, making the topic entity the released answer.
In the source graph, following the same relation chain additionally reaches \texttt{aurangzeb}, \texttt{dara\_shikoh}, and \texttt{murad\_baksh}, all of whom are historically documented siblings of \texttt{shah\_shuja}.
The graph-consistent answer set therefore expands to
\{\texttt{shah\_shuja}, \texttt{aurangzeb}, \texttt{dara\_shikoh}, \texttt{murad\_baksh}\},
while the original released answer and reference path remain unchanged.
A complementary example appears in PQL Question~1064, which asks \emph{``what is the parents of \texttt{Spider-Man\_2}'s award nominations?''}
The released 2-hop annotations follow the relation chain \texttt{award\_nominations} $\rightarrow$ \texttt{parents}, passing through \texttt{Sam\_Raimi} to reach \texttt{Leonard\_Ronald\_Raimi} and \texttt{Celia\_Barbara\_Abrams}.
In the source graph, the same relation chain additionally passes through \texttt{Michael\_Chabon}, reaching his parents, \texttt{Robert\_Chabon} and \texttt{Sharon\_Chabon}.
The graph-consistent answer set thus expands from two released answers to four endpoints, while the number of corresponding path realizations increases from two released references to four source-graph paths.
Unlike the PQ example, where the additional endpoints arise through the same family relation, the PQL expansion introduces an additional intermediate entity that satisfies the released relation chain.
Both examples illustrate that graph restoration can expose additional graph-consistent trajectories and endpoints while preserving the original released answers and reference paths.

\begin{table*}[!t]
    \caption{
    Family-macro structural calibration on the mixed-hop test split.
    The columns distinguish random-walk reachability to the fixed released answer set $A_{\mathrm{released}}$, local navigation-space size, and the multiplicity of graph-consistent answers $A_{\mathrm{graph}}$ and paths $P_{\mathrm{graph}}$.
    RW-Ans$_{\mathrm{MC}}$ reports the mean $\pm$ sample standard deviation over seeds 0, 42, and 100 using 100 walks per question.
    }
    \label{tab:structural_calibration}
    \centering
    \small
    \setlength{\tabcolsep}{5pt}
    \begin{tabular}{llccrrrr}
        \toprule
        & &
        \multicolumn{2}{c}{\textbf{Reachability to $A_{\mathrm{released}}$}} &
        \multicolumn{2}{c}{\textbf{Navigation Space}} &
        \multicolumn{2}{c}{\textbf{Graph-consistent Multiplicity}} \\
        \cmidrule(lr){3-4}
        \cmidrule(lr){5-6}
        \cmidrule(lr){7-8}
        \textbf{Dataset}
        & \textbf{Graph}
        & \textbf{RW-Ans$_p$}
        & \textbf{RW-Ans$_{\mathrm{MC}}$}
        & \shortstack{\textbf{Avg. Actions}}
        & \textbf{Reachability}
        & $\boldsymbol{|A_{\mathrm{graph}}|}$
        & $\boldsymbol{|P_{\mathrm{graph}}|}$ \\
        \midrule
        \multirow{2}{*}{\textsc{PQ}}
        & Filtered & 0.1754 & $0.1750 \pm 0.0010$ & 3.69 & 7.82 & 1.11 & 1.13 \\
        & Source   & 0.0681 & $0.0680 \pm 0.0003$ & 10.70 & 57.41 & 3.17 & 7.48 \\
        \midrule
        \multirow{2}{*}{\textsc{PQL}}
        & Filtered & 0.5395 & $0.5406 \pm 0.0015$ & 3.91 & 4.38 & 1.10 & 1.10 \\
        & Source   & 0.0778 & $0.0776 \pm 0.0007$ & 154.28 & 38,107.71 & 8.41 & 8.66 \\
        \bottomrule
    \end{tabular}
\end{table*}

This expansion has two important consequences for evaluation.
First, graph-relative evaluation can recognize additional endpoints and trajectories that realize the released relation chains rather than treating every deviation from the released realization as an error.
This is particularly important after source restoration, where a policy may follow a trajectory that was unavailable in $G_F$ but is structurally consistent with the same released relation chain in $G_S$.
We therefore retain the released references as the primary fixed target while using graph-consistent answers and paths as complementary measures of behavior that is valid in the active KG.
Appendix~\ref{sec:appendix_graph_path_eval} evaluates the source-graph model trajectories against these graph-consistent path references.

Second, restoration can introduce alternative realizations of released answers that reduce their dependence on cyclic or self-returning navigation.
The released annotations themselves do not change, so their entity revisits and immediate backtracking remain fixed.
However, additional graph structure can make the same released answer reachable through an alternative realization of the annotated relation chain that does not revisit any entity.
This effect is substantial for PQ, where the number of evaluation instances containing at least one cycle-dependent released answer decreases from 2,710 under $G_F$ to 2,100 under $G_S$.
The corresponding change is negligible for PQL, indicating that its released answers remain much more dependent on repeated-entity traversal after source restoration.
Graph restoration can therefore reduce dependence on cyclic navigation without removing the cyclic structure of the original benchmark annotations.

The benchmark-wide prevalence and aggregate magnitude of answer and path expansion are summarized in Table~\ref{tab:reconstruction}, Panel B.
A single QA instance can nevertheless gain as many as 198 graph-consistent paths in PQ and 402 in PQL, showing that the expansion can be highly concentrated.
Additional paths also need not correspond one-to-one with additional answers because multiple restored trajectories can terminate at the same endpoint.
We therefore treat graph-consistent answer multiplicity, path multiplicity, and dependence on cyclic realization as distinct structural properties throughout the audit.

\section{Additional Evaluation Results}
\label{sec:appendix_additional_results}

\subsection{Structural calibration}
\label{sec:appendix_structural_calibration}

Table~\ref{tab:structural_calibration} reports the full family-macro structural calibration underlying the random-walk analysis in Section~\ref{sec:experiments}.
The statistics distinguish three properties of the active graph: unguided reachability to the fixed released answer set $A_{\mathrm{released}}$, the size of the local navigation space encountered by benchmark instances, and the multiplicity of graph-consistent answers and paths induced by the released relation chains.

RW-Ans$_p$ gives the exact probability that uniform random navigation reaches an endpoint in $A_{\mathrm{released}}$, while RW-Ans$_{\mathrm{MC}}$ estimates the same quantity by sampling 100 walks per question for each of three seeds.
Their close agreement across all graph conditions shows that the sampled random-walk reference used in the main results closely tracks the corresponding exact probability.

The two navigation-space columns characterize the local decision environment rather than the global size of the KG.
Avg.\ actions reports the mean number of available navigation actions at states encountered along the released reference trajectories.
Reachability reports the mean number of distinct entities reachable from the topic entity within the fixed navigation horizon.
As in the primary evaluation, this horizon is set to three actions for every question, allowing the filtered and source graphs to be compared under the same navigation budget.
These quantities capture complementary aspects of local search-space expansion: Avg.\ actions measures the action choices exposed at benchmark-relevant states, while Reachability measures the resulting set of distinct entities accessible within the evaluation horizon.

The final two columns report the mean cardinalities of $A_{\mathrm{graph}}$ and $P_{\mathrm{graph}}$.
These sets are obtained by preserving the released relation-chain annotations while evaluating them on the active graph.
They therefore measure how many graph-consistent endpoints and concrete path realizations are available under each graph condition, rather than changing the fixed released targets used by the primary evaluation.

Source restoration changes all three properties in the same direction across PQ and PQL.
The local navigation space expands, additional graph-consistent answers and paths become available, and uniform navigation becomes less likely to terminate at a released answer.
For PQ, the increase in local branching and fixed-horizon reachability is substantial but comparatively moderate.
For PQL, the change is much larger: Avg.\ actions increases from fewer than four to more than 150, while the number of entity reachable within three actions increases from only a few to over 10k.
At the same time, exact random-walk reachability to $A_{\mathrm{released}}$ falls from approximately $54\%$ to below $8\%$.
This provides the structural context for the main-text result that released PQL endpoints are unusually accessible to unguided navigation on the filtered graph but cease to be so after source restoration.

\begin{table}[t]
    \caption{
    Family-macro path fidelity of the answer-oracle shortest path against $P_{\mathrm{released}}$.
    The oracle reaches a released answer for every evaluated question.
    }
    \label{tab:oracle_fidelity}
    \centering
    \small
    \begin{tabular}{llrrrr}
        \toprule
        \textbf{Dataset} & \textbf{Graph} & \textbf{PED}$\downarrow$ & \textbf{RED}$\downarrow$ & \textbf{F1$_{\mathrm{REL}}$}$\uparrow$ & \textbf{F1$_{\mathrm{SG}}$}$\uparrow$ \\
        \midrule
        \multirow{2}{*}{\textsc{PQ}}
        & Filtered & 1.068 & 0.986 & 0.795 & 0.720 \\
        & Source   & 1.772 & 1.381 & 0.686 & 0.465 \\
        \midrule
        \multirow{2}{*}{\textsc{PQL}}
        & Filtered & 1.682 & 1.682 & 0.485 & 0.485 \\
        & Source   & 1.779 & 1.753 & 0.456 & 0.446 \\
        \bottomrule
    \end{tabular}
\end{table}

The increase in $|A_{\mathrm{graph}}|$ and $|P_{\mathrm{graph}}|$ does not mean that released answers or paths are removed.
Rather, restoration preserves the released targets while adding further endpoints and trajectory realizations that are consistent with the same released relation chains.
This distinction motivates the complementary released-target and graph-consistent evaluations reported in Section~\ref{sec:experiments}.

\subsection{Shortest-path oracle fidelity}
\label{sec:appendix_oracle_fidelity}

Table~\ref{tab:oracle_fidelity} reports the complete family-macro path-fidelity metrics for the answer-oracle Shortest Path reference.
The main text retains PED and F1$_{\mathrm{SG}}$ in the primary PQ/PQL result tables, while RED and F1$_{\mathrm{REL}}$ are reported here as supplementary trajectory diagnostics.

\subsection{Graph-consistent trajectory evaluation}
\label{sec:appendix_graph_path_eval}

Table~\ref{tab:graph_consistent_path_eval} evaluates source-graph model trajectories against $P_{\mathrm{graph}}$, complementing the evaluation against the fixed released path references $P_{\mathrm{released}}$ reported in Tables~\ref{tab:pq_policy_results} and~\ref{tab:pql_policy_results}.
The comparison therefore holds the active source graph and predicted trajectories fixed while changing only the path reference from $P_{\mathrm{released}}$ to the larger graph-consistent set $P_{\mathrm{graph}}$.
For MINERVA, the comparison uses the source-dev-selected configuration in both evaluations.

Relative to the source-graph $P_{\mathrm{released}}$ results in Tables~\ref{tab:pq_policy_results} and~\ref{tab:pql_policy_results}, evaluation against $P_{\mathrm{graph}}$ improves trajectory fidelity across all frozen policies and MINERVA.
This recovery shows that some source-graph trajectories that disagree with the particular released path realizations nevertheless follow graph-consistent realizations of the same released relation chains in $G_S$.
The recovery is especially pronounced for PQ, where both PED and F1$_{\mathrm{SG}}$ improve substantially across all evaluated policies.
PQL shows a smaller recovery, particularly for MINERVA, indicating that the expansion from $P_{\mathrm{released}}$ to $P_{\mathrm{graph}}$ explains only part of the trajectory-fidelity degradation observed under source restoration in the main results.
We therefore treat $P_{\mathrm{graph}}$ as a complementary graph-relative diagnostic rather than as a replacement for the fixed released paths used as the primary benchmark target.

\begin{table}[t]
    \caption{
    Source-graph trajectory evaluation against $P_{\mathrm{graph}}$ references for PQ and PQL.
    MINERVA uses the source-dev-selected configuration.
    }
    \label{tab:graph_consistent_path_eval}
    \centering
    \small
    \setlength{\tabcolsep}{3pt}
    \begin{tabular}{llll}
        \toprule
        \textbf{Dataset}
        & \textbf{Method}
        & \textbf{PED}$\downarrow$
        & \textbf{F1$_\mathrm{SG}$}$\uparrow$ \\
        \midrule

        \multirow{4}{*}{PQ}
        & Qwen3 (8B)       & \num{1.5833333333333333} & \num{0.5201269157088122} \\
        & Ministral 3 (8B) & \num{1.2876984126984126} & \num{0.6047713529856387} \\
        & Gemma 4 (E4B)    & \num{1.2392290249433107} & \num{0.6087301587301588} \\
        & MINERVA           & \numstats{1.5346,1.4677,1.5652} & \numstats{0.5227,0.5482,0.5219} \\

        \midrule

        \multirow{4}{*}{PQL}
        & Qwen3 (8B)       & \num{1.9253246753246753} & \num{0.42012987012987013} \\
        & Ministral 3 (8B) & \num{1.6331168831168832} & \num{0.5270562770562771} \\
        & Gemma 4 (E4B)    & \num{1.6525974025974026} & \num{0.5304112554112554} \\
        & MINERVA           & \numstats{1.8636,2.0097,1.9123} & \numstats{0.4575,0.3804,0.4251} \\

        \bottomrule
    \end{tabular}
\end{table}

\section{Relation to Previously Reported Mixed-Hop Results}
\label{app:historical_mixed}

\begin{table*}[t]
    \caption{
    Historical context for mixed-hop PQ/PQL performance (Hits@1).
    Prior values use their respective published mixed-hop splits and preprocessing, whereas MINERVA is evaluated on our family-controlled reconstruction of the filtered KG.
    Values are not directly comparable and are shown only for orientation.
    }
    \label{tab:historical_mixed}
    \centering
    \small
    \begin{tabular}{llll}
        \toprule
        \textbf{Method}
        & \textbf{Setting}
        & \textbf{PQ}$\uparrow$
        & \textbf{PQL}$\uparrow$ \\
        \midrule
        IRN~\cite{zhou2018interpretable}
        & PQ+ / PQL+~\cite{chen2019uhop}
        & \num{0.536}
        & \num{0.529} \\

        HR-BiLSTM + UHop + DQ~\cite{chen2019uhop}
        & PQ+ / PQL+
        & \num{0.9958}
        & \num{0.9183} \\

        ABWIM + UHop + DQ~\cite{chen2019uhop}
        & PQ+ / PQL+
        & \num{0.9944}
        & \num{0.9237} \\

        SRN~\cite{qiu2020stepwise}
        & PQ-M / PQL-M
        & \num{0.893}
        & \num{0.783} \\

        \midrule

        MINERVA
        & Family-controlled, filtered KG
        & \numstats{0.8384,0.8149,0.8329}
        & \numstats{0.9390,0.9484,0.9343} \\
        \bottomrule
    \end{tabular}
\end{table*}

For historical context, Table~\ref{tab:historical_mixed} places evaluation-instance micro Hits@1 for MINERVA on our filtered mixed-hop reconstruction alongside previously reported mixed-hop PQ/PQL results.
We use instance-micro aggregation in this table only to align as closely as possible with the conventional instance-weighted reporting in the earlier literature.
The comparison is not controlled because UHop and SRN construct mixed-hop splits that differ from our family-controlled reconstruction, and their released implementations do not permit us to retrain the corresponding models under our split.
We therefore provide these values only for historical orientation and do not use them to rank methods or support the conclusions of the graph-restoration audit.

The historical methods also differ in what their reported accuracy measures about multi-hop reasoning.
IRN performs iterative relation reasoning rather than explicit graph navigation.
At each hop, it predicts over a global relation vocabulary and uses the predicted relation representation to update a continuous reasoning state.
Although it can also predict intermediate entities, those entities do not become graph states whose outgoing edges define the next action space.
IRN therefore does not execute an entity-to-entity trajectory through the KG.
UHop is closer to graph navigation because its candidate relations are obtained from the current graph context and a selected relation is used to traverse the KG.
However, the model scores relation choices rather than individual outgoing relation--entity edges.
When a selected relation leads to multiple entities, the decision therefore does not identify which destination entity constitutes the next model state.
UHop consequently produces an explicit relation trajectory through the graph, but not necessarily a unique entity-by-entity trajectory of the form evaluated in our navigation setting.
SRN is more directly comparable to MINERVA because its policy selects outgoing relation--entity actions and explicitly updates the current KG entity.
Their PQ-to-PQL differences therefore should not be interpreted as measurements of a common notion of graph-navigation difficulty.

The prior studies also do not isolate a single cause for their lower PQL results.
The original IRN work characterizes PQL as more challenging because it uses a larger KG with fewer training examples, while its broader relation vocabulary also changes the learned relation-prediction problem.
UHop reports a different relation-candidate distribution on PQL and finds that most PQ/PQL errors arise from relation extraction rather than termination.
SRN directly navigates outgoing KG actions, but its experiments do not separate the effects of graph structure, relation representation, training-data availability, and beam-search behavior.
We therefore treat the historical PQ-to-PQL decline as an empirical property of those published model--benchmark configurations rather than evidence that PQL is intrinsically harder to navigate.

Our filtered reconstruction exhibits the opposite endpoint-accuracy trend.
MINERVA obtains higher Hits@1 on PQL than on PQ, and the same direction appears for all three frozen SLM policies and the random-walk calibration in Tables~\ref{tab:pq_policy_results} and~\ref{tab:pql_policy_results}.
For example, RW-Ans$_{\mathrm{MC}}$ rises from $0.175$ on PQ to $0.541$ on PQL.
Because the historical results differ from our reconstruction in splits, preprocessing, training objectives, and evaluation protocols, this reversal cannot be attributed specifically to model architecture.
Instead, it shows that benchmark name and nominal graph scale alone do not determine effective navigation difficulty.
The instantiated graph and evaluation protocol can materially change the task presented to a navigation policy.

\section{Cross-graph Transfer of Learned Navigation}
\label{sec:appendix_cross_graph_transfer}

\begin{table}[t]
    \caption{
Cross-graph transfer of MINERVA on PQ under a shared entity/relation vocabulary.
Each checkpoint is tuned on the development split of its training KG and evaluated on both graph conditions without retuning.
$G_F \rightarrow G_S$ and $G_S \rightarrow G_F$ therefore measure graph-expansion and graph-contraction transfer, respectively.
}
    \label{tab:pq_minerva_transfer}
    \centering
    \small
    \setlength{\tabcolsep}{2pt}
    \begin{tabular}{clll}
        \toprule
        \textbf{Train $\rightarrow$ Eval}
        & \textbf{H@1}$\uparrow$
        & \textbf{PED}$\downarrow$
        & \textbf{F1$_{\mathrm{SG}}$}$\uparrow$ \\
        \midrule

        $G_F \rightarrow G_F$
        & \numstats{0.8450,0.8209,0.8393}
        & \numstats{0.4524,0.5320,0.5748}
        & \numstats{0.8617,0.8382,0.8228} \\

        $G_F \rightarrow G_S$
        & \numstats{0.4110,0.4328,0.4481}
        & \numstats{1.7168,1.6973,1.6831}
        & \numstats{0.4247,0.4337,0.4313} \\

        \addlinespace

        $G_S \rightarrow G_S$
        & \numstats{0.5592,0.5456,0.5760}
        & \numstats{1.8827,1.8886,1.8716}
        & \numstats{0.3906,0.3898,0.3980} \\

        $G_S \rightarrow G_F$
        & \numstats{0.5505,0.5830,0.5836}
        & \numstats{1.2664,1.1893,1.2942}
        & \numstats{0.6314,0.6486,0.6319} \\

        \bottomrule
    \end{tabular}
\end{table}

The primary MINERVA comparison in Sec.~\ref{sec:experiments} trains and evaluates each policy on the same active KG.
A complementary question is whether a policy trained under one graph condition transfers to the other without additional training or destination-graph tuning.
In particular, does the learned policy remain robust when the graph structure changes, or does it become specialized to its training environment even though the same released answers and annotated reference paths remain available in both KGs?
As in Tables~\ref{tab:pq_policy_results} and~\ref{tab:pql_policy_results}, the model entity and relation ID vocabularies are fixed across both graph conditions, so cross-graph evaluation does not introduce an out-of-vocabulary mismatch.
The transfer intervention can nevertheless expose graph structure that was absent during training, particularly when a policy trained on the filtered graph is evaluated on the restored source graph.

For each training condition, hyperparameters are selected on the corresponding development graph and the resulting checkpoints are evaluated on both $G_F$ and $G_S$.
Thus, $G_F \rightarrow G_S$ measures transfer under graph expansion, whereas $G_S \rightarrow G_F$ measures transfer under graph contraction.
The same trained seeds are reused across the paired evaluation conditions, so the comparison changes only the graph presented to a fixed trained policy.

Tables~\ref{tab:pq_minerva_transfer} and~\ref{tab:pql_minerva_transfer} show a consistent directional asymmetry across both benchmarks.
Policies trained on $G_F$ transfer poorly to $G_S$, with substantial degradation in both answer accuracy and trajectory fidelity.
In contrast, when policies trained on $G_S$ are transferred to $G_F$, their answer accuracy remains comparable to or improves over their in-domain $G_S \rightarrow G_S$ performance, while their path fidelity improves substantially relative to the same source-graph baseline.
The graph-expansion penalty is particularly pronounced for PQL, consistent with the substantially larger change in its effective navigation search space under source restoration.
Because the released answers and reference paths remain available in both graph conditions, the degradation under graph expansion cannot be attributed to the target trajectory disappearing.
Instead, the results indicate that successful navigation on $G_F$ depends in part on the surrounding alternatives encountered during training: when additional graph structure is restored, the learned action preferences do not remain invariant.
The reverse direction provides a useful control, since removing alternatives from a policy trained on $G_S$ does not produce the same loss and instead tends to improve agreement with the released reference paths.
The transfer asymmetry therefore isolates sensitivity to the graph environment itself, strengthening the audit conclusion that graph construction can change the learned navigation problem even when the nominal QA targets are held fixed.
We treat this asymmetry as a transfer diagnostic for MINERVA rather than as a general claim that source-graph training necessarily improves cross-graph robustness for other navigation architectures.

The present experiment evaluates only the two endpoint environments, $G_F$ and $G_S$.
A useful follow-up would construct nested intermediate graphs satisfying
$G_F \subset G_i \subset G_S$
and progressively restore the triples removed by benchmark filtering.
This would allow the transfer asymmetry to be traced as a function of graph restoration, helping distinguish a gradual sensitivity to increasing structural alternatives from effects driven by particular subsets of restored triples.

\begin{table}[t]
    \caption{
Cross-graph transfer of MINERVA on PQL under a shared entity/relation vocabulary.
Each checkpoint is tuned on the development split of its training KG and evaluated on both graph conditions without retuning.
$G_F \rightarrow G_S$ and $G_S \rightarrow G_F$ measure expansion and contraction transfer, respectively.
}
    \label{tab:pql_minerva_transfer}
    \centering
    \small
    \setlength{\tabcolsep}{2pt}
    \begin{tabular}{clll}
        \toprule
        \textbf{Train $\rightarrow$ Eval}
        & \textbf{H@1}$\uparrow$
        & \textbf{PED}$\downarrow$
        & \textbf{F1$_{\mathrm{SG}}$}$\uparrow$ \\
        \midrule

        $G_F \rightarrow G_F$
        & \numstats{0.9416,0.9448,0.9351} 
        & \numstats{1.2857,1.2662,1.2955} 
        & \numstats{0.7353,0.7255,0.7179} \\

        $G_F \rightarrow G_S$
        & \numstats{0.2662,0.3084,0.4026}
        & \numstats{2.3571,2.2857,2.3571} 
        & \numstats{0.2108,0.2371,0.2329} \\

        \addlinespace

        $G_S \rightarrow G_S$
        & \numstats{0.8377,0.8149,0.8539}
        & \numstats{1.8831,2.0357,1.9448}
        & \numstats{0.4419,0.3536,0.4119} \\

        $G_S \rightarrow G_F$
        & \numstats{0.8636,0.8506,0.9026}
        & \numstats{1.3019,1.5584,1.4643}
        & \numstats{0.7218,0.6225,0.6764} \\

        \bottomrule
    \end{tabular}
\end{table}

\section{Preliminary Wording Audit}
\label{sec:appendix_wording}

During reconstruction, we observed cases in which the generated natural-language question did not faithfully reflect the semantics of its released reasoning path.
To assess whether these examples were isolated, we conducted an exploratory LLM-assisted audit of the released \texttt{PQ-2H.txt}, \texttt{PQ-3H.txt}, \texttt{PQL-2H.txt}, and \texttt{PQL-3H.txt} files alongside our reconstructed QA representations, using OpenAI ChatGPT (GPT-5.6 Sol, accessed September 30, 2026).
The model was instructed to preserve the original benchmark annotations rather than repair them, distinguish explicit semantic contradictions from unsupported lexical restrictions, and examine released reasoning paths and cross-question evidence whenever possible.
The prompt and outputs are provided as artifacts of this audit.
Because the audit is LLM-assisted and does not constitute exhaustive human annotation, we do not interpret its aggregate flag rates as benchmark error rates or use them as evidence for the structural claims in the main paper.
Furthermore, some classifications, particularly semantic underspecification and type-role anomalies, remain heuristic.
Nevertheless, several directly inspectable examples reveal recurring inconsistencies between the natural-language wording and the annotated reasoning structure.

\paragraph{PQ: gender-specific template drift.}
PQ uses generic kinship relations, including \texttt{children}, \texttt{parents}, and \texttt{spouse}, but the corresponding questions sometimes replace these relations with gender-specific terms such as \emph{daughter}, \emph{dad}, or \emph{husband}.
Such lexicalizations impose semantic restrictions not encoded by the generic relation and can directly contradict the released annotations.
For example, the Question 15 \emph{``the sex of \texttt{claudius}'s husband?''} follows the annotated path
\texttt{claudius} $\xrightarrow{\texttt{spouse}}$ \texttt{aelia\_paetina} $\xrightarrow{\texttt{gender}}$ \texttt{female}.
The term \emph{husband} presupposes a male spouse, whereas the annotated path explicitly identifies the intermediate entity as female.
Similarly, Question 60 \emph{``what is the \texttt{princess\_elizabeth\_of\_england}'s dad's sex?''} follows
\texttt{princess\_elizabeth\_of\_england} $\xrightarrow{\texttt{parents}}$ \texttt{henrietta\_maria\_of\_france} $\xrightarrow{\texttt{gender}}$ \texttt{female}.
Here, the generic \texttt{parents} relation is rendered as \emph{dad}, despite the annotated intermediate entity being female.
Both contradictions are directly observable within their respective released reasoning paths, without requiring external biographical verification.

Other inconsistencies become apparent when comparing annotations across released questions.
For example, Question 429 \emph{``the children of daughter of \texttt{p\_j\_kennedy}?''} follows
\texttt{p\_j\_kennedy} $\xrightarrow{\texttt{children}}$ \texttt{joseph\_p\_kennedy\_sr} $\xrightarrow{\texttt{children}}$ \texttt{rosemary\_kennedy}.
Although the question describes the intermediate entity as a \emph{daughter}, other released PQ annotations explicitly identify \texttt{joseph\_p\_kennedy\_sr} as male.
These examples indicate that the surface wording can introduce constraints absent from, or inconsistent with, the released graph annotations.
They do not imply that every gender-specific lexicalization is incorrect, since some may agree with the corresponding entities even when the underlying relation remains generic.

\paragraph{PQL: predicate lexicalization collapse.}
PQL exhibits a different, mechanically identifiable phenomenon: distinct Freebase-style predicates can share the same visible relation phrase.
In the inspected released annotations, 125 distinct full predicates are represented by 111 surface relation phrases, with 10 phrases corresponding to more than one predicate.
For example, \emph{genre} represents five distinct predicates spanning books, broadcast content, films, music albums, and television programs.
Likewise, \emph{releases} represents the distinct predicates
\texttt{\_\_music\_\_album\_\_releases},
\texttt{\_\_music\_\_record\_label\_\_releases}, and
\texttt{\_\_music\_\_recording\_\_releases}.
Consequently, the generated question does not always preserve the predicate identity encoded in its released relation chain.

A concrete example is PQL Question~1663:
\emph{``what is the releases of releases of \texttt{Amsterdam}'s artist?''}
The reconstructed instance contains two released relation chains,
\texttt{r2272}$\rightarrow$\texttt{r2204}$\rightarrow$\texttt{r2279}
and
\texttt{r2272}$\rightarrow$\texttt{r2204}$\rightarrow$\texttt{r2204},
which differ in their final predicate.
These final predicates correspond to
\texttt{\_\_music\_\_recording\_\_releases}
and
\texttt{\_\_music\_\_album\_\_releases},
respectively, but both are rendered as \emph{releases}.
The released answers include \texttt{The\_Muffs} and \texttt{Another\_Day}.
All eight reconstructed PQL instances containing multiple released relation chains for the same question and answer set exhibit this particular final-hop lexicalization collision.
Thus, these cases demonstrate a reproducible loss of relation-specific information in the question wording rather than merely awkward grammatical construction.

\paragraph{PQL: semantic type and argument-role mismatches.}
The exploratory audit also identifies compositions in which successive predicates imply incompatible semantic roles.
For example, the released question
\emph{``what is the tracks of \texttt{Alfred\_Philippe\_Roll}'s place\_of\_death?''}
follows the relation chain
\texttt{\_\_people\_\_deceased\_person\_\_place\_of\_death}
$\rightarrow$
\texttt{\_\_music\_\_recording\_\_tracks},
with the intermediate entity \texttt{Paris}.
The first predicate identifies a location, whereas the second belongs to a music-recording namespace and ordinarily expects a different source-entity role.
This discrepancy is observable from the released relation names and their composition, although it does not by itself establish that the underlying KG edges are invalid.
We therefore treat such cases as possible semantic mismatches in the graph-to-language interface, rather than independently verified factual errors.

\paragraph{Scope and implications.}
These findings identify a separate source of ambiguity in PQ/PQL: the correspondence between natural-language questions and their annotated reasoning structures.
In PQ, gender-specific lexicalizations can impose constraints that are absent from, or contradict, the underlying relations and entity annotations.
In PQL, predicate lexicalization can obscure distinctions between relation types, while some generated questions combine predicates with potentially incompatible semantic roles.
For question-conditioned navigation, these inconsistencies matter because the question is expected to guide relation selection, yet its wording may not uniquely or faithfully specify the annotated relation chain.
Consequently, a trajectory that follows the released annotations need not fully satisfy the natural-language question, while a linguistically plausible interpretation may disagree with the released reference.

This distinction also qualifies the interpretation of graph-consistent evaluation.
Our graph-relative answer and path sets are defined by realizations of the released relation chains, not by independent semantic validation of the natural-language questions.
Thus, an endpoint or trajectory can be structurally consistent with the annotated chain without necessarily constituting a semantically correct answer to the question as written.
Although the exploratory audit quantifies the prevalence of flagged wording inconsistencies under its classification criteria, these flags remain to be independently validated through systematic human annotation.
Consequently, the reported counts should not be interpreted as confirmed benchmark error rates, and the audit does not establish the causal contribution of these inconsistencies to navigation accuracy or trajectory fidelity.
Importantly, the original question wording and annotations are held fixed across our filtered and source graph conditions.
We therefore preserve the released questions and annotations rather than repairing or excluding flagged instances, maintaining the matched intervention that isolates the effect of graph construction.
The wording findings remain auxiliary to this comparison and motivate future human validation of question--relation alignment, while the mechanically verifiable lexicalization collisions provide reproducible evidence of information loss in the generated questions.

\section{Cautions for Interpreting PQ/PQL Results}
\label{app:implications}

PQ/PQL remain useful controlled benchmarks, but several properties of their released artifacts limit the conclusions that can be drawn from performance on them.
This appendix summarizes those interpretive boundaries rather than restating the empirical comparisons reported in Section~\ref{sec:experiments}.

\paragraph{Filtered-graph results do not establish source-KG performance.}
A result obtained on the released task-filtered graph should be interpreted as performance within that particular navigation environment.
It should not, without additional evidence, be extrapolated to navigation or retrieval over the larger source KG from which the benchmark was derived.
For retrieval-based systems, this distinction is especially important because the released graph already restricts which entities, relations, and edges are available before retrieval begins.
Accordingly, strong performance on PQ/PQL does not by itself demonstrate robustness to substantially larger candidate spaces or denser graph neighborhoods.

\paragraph{Graph consistency does not establish language-level correctness.}
Graph-consistent answers and paths certify consistency with the released relation chains under the active graph, not independent semantic correctness of the natural-language question.
As Appendix~\ref{sec:appendix_wording} illustrates, mechanically generated PQ/PQL questions can express constraints or ambiguities that are not captured by those structural annotations.
Released answers, graph-consistent answers, and answers justified by the full surface semantics should therefore not be treated as interchangeable notions of correctness.

\paragraph{Answer accuracy does not establish faithful reasoning.}
A correct endpoint should not be interpreted as evidence that the released reasoning trajectory was followed.
Conversely, disagreement with one released path does not establish that a trajectory is structurally invalid when another graph-consistent realization exists.
Endpoint correctness and trajectory agreement should therefore be evaluated separately when making claims about navigation or reasoning behavior.

\paragraph{Nominal hop count does not necessarily measure required reasoning depth.}
The annotated hop count describes the released reference walk rather than the minimum number of actions required to reach a valid answer.
As Appendix~\ref{sec:appendix_topic_answer} shows, cyclic or self-returning references and topic--answer identity can admit shorter correct routes, including zero-action solutions.
Claims about reasoning depth should therefore not be based on the nominal 2-hop or 3-hop label alone.

\paragraph{The benchmark name does not uniquely specify the experimental setting.}
Published PQ/PQL results may use different hop-specific or mixed-hop reconstructions, graph artifacts, action definitions, stopping conventions, inverse relations, candidate limits, answer aggregation rules, and path-reference handling.
Trainable navigation systems can additionally differ in which graph condition is used for training and hyperparameter selection.
Consequently, results labeled simply as ``PQ'' or ``PQL'' should not be assumed directly comparable unless these choices are sufficiently aligned.
Studies using these benchmarks should identify the graph artifact and preprocessing used and state how multiple answers, multiple paths, repeated entities, stopping actions, and aggregation are handled.

\paragraph{Scope of interpretation.}
These cautions do not invalidate results obtained on the released PQ/PQL benchmarks.
They instead delimit what those results directly support.
PQ/PQL remain useful controlled benchmarks for studying relation composition and graph navigation when their graph, annotation, and language conventions are made explicit.
Claims about source-KG robustness, retrieval difficulty, unique reasoning explanations, required reasoning depth, or fine-grained natural-language understanding require additional evidence beyond performance on the released benchmark alone.
%


\end{document}